\documentclass{article}

\usepackage[preprint]{neurips_2026}

\usepackage[utf8]{inputenc} 
\usepackage[T1]{fontenc}    
\usepackage{hyperref}       
\usepackage{url}            
\usepackage{booktabs}       
\usepackage{amsfonts}       
\usepackage{nicefrac}       
\usepackage{microtype}      
\usepackage[table]{xcolor}  

\usepackage{graphicx}
\usepackage{subcaption}
\usepackage{amsmath}
\usepackage{amssymb}
\usepackage{multirow}
\usepackage{bbm}
\usepackage{algorithm}
\usepackage{algorithmicx}
\usepackage{stmaryrd}
\usepackage[noend]{algpseudocode}
\usepackage{makecell}
\usepackage{wrapfig}

\title{Learning Human-Intention Priors from Large-Scale Human Demonstrations for Robotic Manipulation}

\author{
    \vspace{-9mm} \\
    \textbf{
    Yifan Xie$^{1,2\ast}$
    \quad YuAn Wang$^{2}$\thanks{: Equal contribution;\quad $\dagger$: Project lead;\quad $\ddagger$: Corresponding author.}
    \quad Guangyu Chen$^{1}$
    \quad Jinkun Liu$^{1}$
    } \vspace{1mm} \\
    \textbf{
    Yu Sun$^{2\dagger}$
    \quad Wenbo Ding$^{1\ddagger}$
    } \vspace{2mm} \\
    $^1$Tsinghua University\quad $^2$ByteDance \vspace{1mm}\\
}

\begin{document}

\maketitle

\begin{abstract}
Human videos contain rich manipulation priors, but using them for robot learning remains difficult because raw observations entangle scene understanding, human motion, and embodiment-specific action. We introduce MoT-HRA, a hierarchical vision-language-action framework that learns human-intention priors from large-scale human demonstrations. We first curate HA-2.2M, a 2.2M-episode action-language dataset reconstructed from heterogeneous human videos through hand-centric filtering, spatial reconstruction, temporal segmentation, and language alignment. On top of this dataset, MoT-HRA factorizes manipulation into three coupled experts: a vision-language expert predicts an embodiment-agnostic 3D trajectory, an intention expert models MANO-style hand motion as a latent human-motion prior, and a fine expert maps the intention-aware representation to robot action chunks. A shared-attention trunk and read-only key-value transfer allow downstream control to use human priors while limiting interference with upstream representations. Experiments on hand motion generation, simulated manipulation, and real-world robot tasks show that MoT-HRA improves motion plausibility and robust control under distribution shift.
\end{abstract}

\section{Introduction}

Generalist robotic manipulation is increasingly framed as vision-language-action (VLA) modeling, where a single policy conditions on visual observations and language instructions and produces executable actions~\cite{zitkovich2023rt,o2024open,mees2024octo,kim2025openvla,black2024pi_0}. This paradigm has benefited from large pretrained vision-language backbones and growing cross-embodiment robot datasets, but its scaling path is still constrained by the cost and sparsity of robot demonstrations. Robot trajectories are expensive to collect, tied to specific hardware, and often cover only a narrow slice of everyday manipulation. In contrast, human videos record abundant object interactions across homes, kitchens, workshops, viewpoints, and task styles~\cite{miech2019howto100m,grauman2022ego4d,damen2020epic,goyal2017something}. If these videos can be converted into supervision for control, they offer a much broader source of manipulation priors than robot-only data.

However, human videos are not robot demonstrations. Raw clips entangle scene understanding, hand motion, camera motion, task progress, and embodiment-specific constraints. Many segments contain visible hands without purposeful manipulation, while useful interaction clips rarely provide temporally aligned action labels or robot-executable controls. Recent human-centric robot learning methods have begun to exploit this resource by extracting hand motion, aligning visual observations with physical space, or introducing latent bridges between videos and actions~\cite{luo2025being,li2025scalable,feng2025spatial,beingbeyond2026beingh07,chen2025flowing}. These efforts suggest that human videos contain transferable physical knowledge, but they also expose a core representation problem. To benefit robot learning, a model must separate \emph{where} an interaction should occur, \emph{how} human motion expresses task intent, and \emph{how} a particular robot should execute the task. 
Collapsing these factors into a single action representation can cause downstream robot loss to overwrite useful human-motion priors, and can make the learned policy brittle when visual appearance, viewpoint, object instance, or embodiment changes.

Our key observation is that human demonstrations are most useful when they are treated not as noisy substitutes for robot actions, but as structured evidence about manipulation intent. 
A human hand trajectory provides an embodiment-agnostic spatial scaffold, and articulated hand motion captures temporal coordination and contact-oriented motor preference. Robot controls should then be learned as an embodiment-specific realization of this prior.
This view suggests a hierarchical representation for human-to-robot transfer: first ground the interaction in 3D space, then model human intention in a structured hand-motion space, and finally adapt the intention-aware representation to robot action chunks. Such a decomposition preserves the reusable parts of human behavior while leaving the final policy free to match the kinematics and action convention of the target robot.

To realize this idea, we present MoT-HRA, a hierarchical VLA framework that learns human-intention priors from large-scale human demonstrations and transfers them to robotic manipulation. We first curate HA-2.2M, a 2.2M-episode action-language dataset reconstructed from heterogeneous human videos through hand-centric filtering, spatial reconstruction, temporal segmentation, and language alignment. Building on this dataset, MoT-HRA factorizes action generation into three coupled experts: a vision-language expert that predicts an embodiment-agnostic 3D trajectory, an intention expert that models a MANO-style human hand-motion prior, and a fine expert that maps the intention-aware representation to robot action chunks. A shared-attention trunk supplies common multimodal context, while read-only key-value transfer implements knowledge insulation~\cite{driess2025knowledge}, allowing downstream robot control to use human priors without overwriting upstream spatial and intention representations. This design turns heterogeneous human videos into an intermediate intention manifold rather than forcing them into robot-specific action labels.

Our main contributions can be summarized as follows:
\begin{itemize}
    \item We introduce HA-2.2M, a large-scale human-demonstration dataset built from heterogeneous web and egocentric videos. It provides temporally coherent action-language episodes with reconstructed hand-centric spatial supervision, offering a scalable source of manipulation priors beyond robot-only demonstrations.
    \item We propose MoT-HRA, a hierarchical Mixture-of-Transformer VLA architecture that separates spatial trajectory grounding, latent human-intention modeling, and embodiment-specific robot action generation, with knowledge insulation to reduce destructive interference between human-prior learning and robot policy learning.
    \item We validate the proposed decomposition on hand motion generation, SimplerEnv, and real-world manipulation tasks, showing that human-intention priors improve motion plausibility, spatial grounding, and control robustness under distribution shift.
\end{itemize}

\section{Related Work}
\subsection{Vision-Language-Action Models}
Vision-language-action (VLA) models extend pretrained multimodal representations from perception and instruction following to robot control. Representative systems ground visual-language knowledge in robot action datasets and language-conditioned policies~\cite{zitkovich2023rt,o2024open,mees2024octo,kim2025openvla}, while recent work improves action generation through flow-based policies, efficient action tokenization, fine-tuning recipes, spatial representations, latent planning, and world-model interfaces~\cite{black2024pi_0,pertsch2025fast,kim2025fine,li2026matters,qu2025spatialvla,huang2025thinkact,team2025gigabrain}. Other studies focus on preserving pretrained knowledge and strengthening multimodal or world model representations for control~\cite{driess2025knowledge,beingbeyond2026beingh07,xie2025universal,li2026causal,ye2026world}. Our work follows this factorized trend by separating visual-language grounding, human-intention modeling, and robot-specific action prediction.


\subsection{Human Demonstration Learning}
Human demonstration learning studies how passive recordings can supervise robot policies without robot actions. Human and egocentric videos have been used to learn visual representations, rewards, and affordances that expose object-centric goals, contact cues, and task progress~\cite{nair2022r3m,ma2023vip,bahl2023affordances,mees2023grounding}. Recent human-video-to-VLA work frames these signals as representation bridges from unstructured videos to executable actions~\cite{feng2026fromhumanvideos}. Large corpora such as Ego4D, EPIC-KITCHENS, Something-Something-V2, and Ego-Exo4D provide diverse interactions~\cite{grauman2022ego4d,damen2020epic,goyal2017something,grauman2024egoexo4d}, but usually lack aligned control labels, calibrated 3D state, and proprioception.

Existing methods differ in the bridge they recover. Latent-action methods tokenize inter-frame changes~\cite{ye2025latent,chen2024igor,chen2025moto,bu2025univla}, world-model methods transfer future-prediction or generated-video priors to action heads~\cite{wu2024unleashing,cheang2024gr2,bharadhwaj2025gen2act}, and explicit methods use 2D tracks or affordance prompts~\cite{wen2024anypoint,yang2025magma} or reconstruct 3D wrist, hand, object, and MANO-style motion~\cite{qin2022dexmv,chen2025vidbot,yang2025egovla,luo2025being,li2025scalable,chen2025flowing}. Our method follows the explicit 3D line but separates reconstructed human hand motion as an intention prior from the final embodiment-specific action expert, reducing interference when transferring human demonstrations to robot control.


\section{Methods}

In this section, we describe the full pipeline for learning robot control from large-scale human demonstrations. We first introduce HA-2.2M (Sec.~\ref{HA-2.2M}), a 2.2M-episode dataset with aligned action and language supervision reconstructed from raw human videos. We then present MoT-HRA (Sec.~\ref{MoT-HRA}), a hierarchical Mixture-of-Transformers architecture that converts these demonstrations into a latent intention space connecting multimodal understanding, human motion priors, and embodiment-specific action generation.

\begin{figure}
    \centering
    \includegraphics[width=\linewidth]{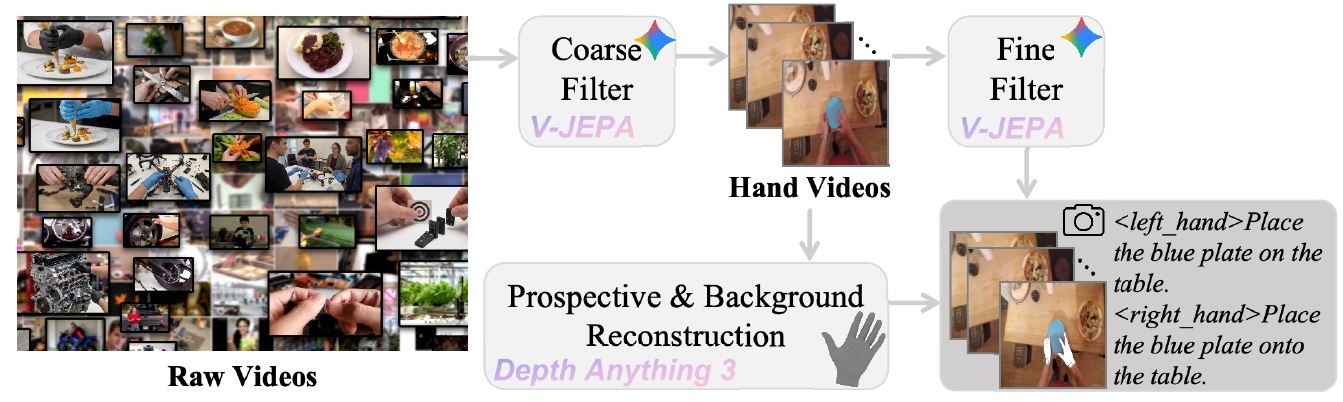}
    \caption{
Overview of the HA-2.2M curation pipeline. Large-scale unlabeled human demonstration videos are converted into 2.2M action-language episodes through coarse filtering, perspective and background reconstruction, and fine filtering. These stages retain manipulation-relevant hand videos, recover image-aligned hand motion and scene context, and extract temporally coherent atomic episodes with aligned action descriptions. The pie chart summarizes the source composition of the final dataset.
    }
    \label{fig:data_pipeline}
\end{figure}

\subsection{HA-2.2M} \label{HA-2.2M}
\paragraph{Overview}
HA-2.2M contains 2.2M short human-demonstration episodes from egocentric and third-person views~\cite{miech2019howto100m,grauman2022ego4d,damen2020epic,goyal2017something}.
Rather than reuse heterogeneous source annotations, we reconstruct action-language episodes from raw videos with the pipeline in Fig.~\ref{fig:data_pipeline}. This mixture combines egocentric recordings with diverse third-person videos, broadening coverage over objects, scenes, and manipulation styles. At the same time, it introduces noise from passive hand motion, abrupt viewpoint changes, camera motion unrelated to interaction, and mismatched temporal granularity. The final pipeline applies coarse filtering, spatial reconstruction, and fine filtering so that each retained clip corresponds to a short manipulation primitive. These clips serve as pretraining units for learning latent human action priors instead of merely enlarging the visual dataset.

\paragraph{Coarse Filter}
Raw web videos often show hands without purposeful manipulation, such as idle gestures, transitional motion, or camera motion unrelated to interaction. We therefore use a recall-oriented cascade: Gemini~\cite{team2023gemini} first labels likely hand-centric action segments, and a lightweight V-JEPA-based~\cite{assran2025v} classifier is then trained on these labels and applied across the full dataset. The first stage removes obviously irrelevant content while preserving diverse manipulation patterns across viewpoints and environments, and the second suppresses visible-hand false positives introduced by multimodal labeling. This design avoids expensive multimodal calls for every segment while retaining the breadth needed for large-scale pretraining, yielding a cleaner input distribution for the reconstruction stage.


\paragraph{Perspective and Background Reconstruction}
The retained videos are converted into image-aligned 3D trajectories by reconstructing hand motion and scene context. For each frame, we localize hands with VitPose~\cite{xu2022vitpose,xu2023vitpose++}, refine the boxes for robustness under blur and occlusion, and estimate absolute-scale MANO hand pose with HaMeR~\cite{pavlakos2024reconstructing}. Since HaMeR outputs crop-space predictions, we recover crop-to-image translation and scale, map MANO states back to the original frame, and temporally smooth trajectories to reduce jitter. In parallel, Depth Anything 3~\cite{lin2025depth} predicts relative monocular depth for background layout, which we align to the hand's absolute scale by fitting DA3 values on hand pixels to the MANO reconstruction. This places scene geometry in a hand-consistent 3D frame and gives subsequent segmentation and language alignment access to scaled scene context rather than appearance-only videos.

\paragraph{Fine Filter}
After reconstruction, we extract atomic episodes and align them with language supervision. A V-JEPA~\cite{assran2025v} temporal segmentation model, trained on 38K manually annotated segments, predicts action boundaries so that clips correspond to manipulation units rather than arbitrary fixed windows. Because the high-recall boundary predictor can over-segment complex activities, Gemini~\cite{team2023gemini} merges adjacent clips with continuous intent and generates concise action descriptions, including hand-specific tags when the two hands play different roles. The detailed merge rules are deferred to the supplementary material. This final stage turns noisy reconstructed trajectories into temporally coherent action-language episodes for HA-2.2M.

\begin{figure}
    \centering
    \includegraphics[width=0.95\linewidth]{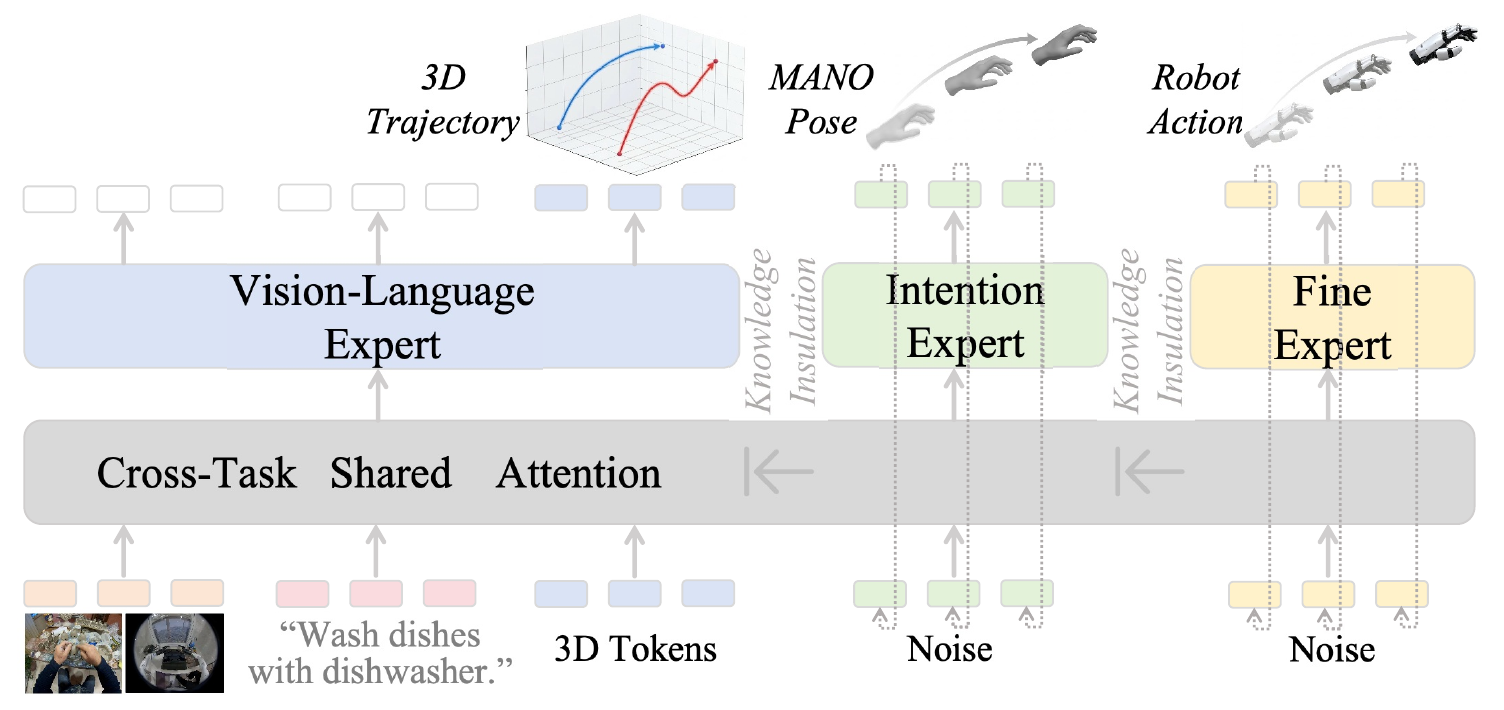}
    \caption{
Overview of MoT-HRA. 
Given an image, a language instruction, and chunk-sized query tokens, the model predicts robot actions with three hierarchical experts: a vision-language expert that autoregressively predicts an embodiment-agnostic discretized 3D trajectory, 
an intention expert that generates a MANO-style hand-pose sequence by denoising noise conditioned on the spatial plan,
and a fine expert that maps the resulting intention-aware representation to embodiment-specific robot actions. A shared-attention trunk supplies common multimodal context, while read-only key-value transfer across experts enforces knowledge insulation and preserves human motion priors during downstream action learning.
}
    \label{fig:method}
\end{figure}

\subsection{MoT-HRA} \label{MoT-HRA}

\paragraph{Problem Formulation}
Large-scale human videos provide useful priors for robotic control, but direct observation-to-action learning entangles visual grounding, human motion intent, and embodiment-specific actuation~\cite{luo2025being,li2025scalable,feng2025spatial}. 
A single representation must absorb both transferable human manipulation intention and low-level robot actions. 
MoT-HRA instead factorizes the chunk policy through an embodiment-agnostic spatial plan and a learned human-intention state. Given an observation image $o$, a language instruction $w$, and a chunk horizon $H$, we approximate
\begin{equation}
\begin{aligned}
p_{\theta}(a_{1:H}\mid o,w)
&\approx
\int
p_{\theta_a}(a_{1:H}\mid o,w,\tau_{1:H},h^{\mathrm{int}}_{1:H})\,
q_{\theta_m}(h^{\mathrm{int}}_{1:H}\mid o,w,\tau_{1:H})
p_{\theta_{\tau}}(\tau_{1:H}\mid o,w)
\,\mathrm{d}\tau\,\mathrm{d}h .
\end{aligned}
\end{equation}
Here $\tau_{1:H}$ is an embodiment-agnostic 3D plan, and $h^{\mathrm{int}}_{1:H}$ is the latent intention representation passed to the robot-action factor. The MANO-style pose sequence $m_{1:H}$ supervises the intention branch but is not treated as a robot command. In implementation, the factors are sampled sequentially:
\begin{equation}
\begin{aligned}
\tau_{1:H} &\sim p_{\theta_{\tau}}(\cdot \mid o,w), \\
(m_{1:H},h^{\mathrm{int}}_{1:H})
&= \Phi_{\theta_m}(\epsilon^m; o,w,\tau_{1:H}), \\
a_{1:H}
&= \Phi_{\theta_a}(\epsilon^a; o,w,\tau_{1:H},h^{\mathrm{int}}_{1:H}),
\end{aligned}
\end{equation}
where $\epsilon^m,\epsilon^a\sim\mathcal{N}(0,I)$, and $\Phi_{\theta_m}$ and $\Phi_{\theta_a}$ denote the conditional flow-matching solution maps of the intention and fine experts. The action flow conditions on $h^{\mathrm{int}}_{1:H}$ rather than treating generated hand poses as robot commands. Thus, the 3D trajectory specifies where the interaction should unfold, the intention branch captures human motion regularities, and the final expert adapts this bridge to the target embodiment.

\begin{wrapfigure}{r}{0.45\textwidth}
    \centering
    \includegraphics[width=\linewidth]{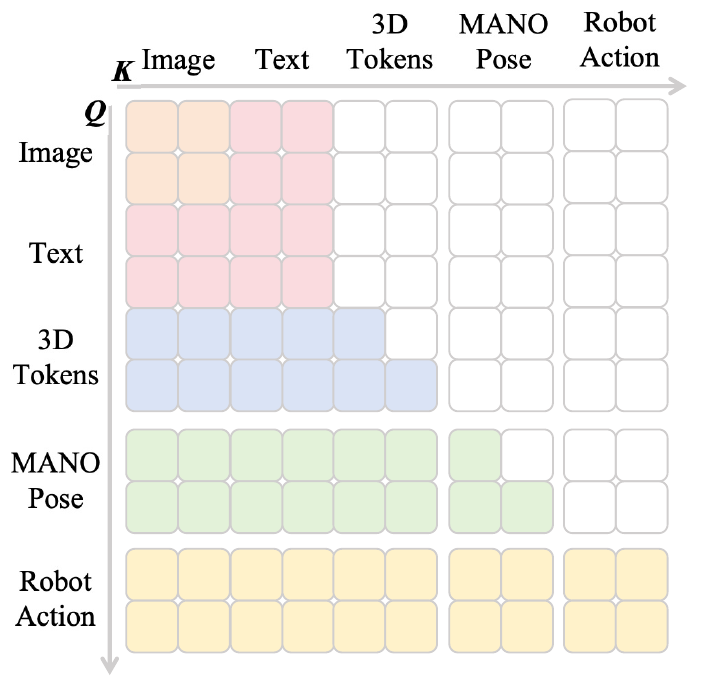}
    \caption{Attention mask of MoT-HRA. Image and text tokens use bidirectional attention to build a shared multimodal context, while 3D trajectory and MANO pose tokens attend to the full prefix but remain causally masked within their own spans. Robot-action tokens attend to all preceding modalities and use bidirectional attention within the action chunk, enabling joint refinement of temporally coupled controls.
}
    \label{fig:attention}
\end{wrapfigure}

\paragraph{Unified Architecture}
As shown in Fig.~\ref{fig:method}, MoT-HRA instantiates this factorization with a Mixture-of-Transformers architecture composed of a cross-task shared-attention trunk and three task-specific experts: a vision-language expert, an intention expert, and a fine expert. The full token sequence is
\begin{equation}
X =
\big[
x^{\text{img}}_{1:N_v};
x^{\text{txt}}_{1:N_l};
z^{3d}_{1:H};
z^{m}_{1:H};
z^{a}_{1:H}
\big],
\end{equation}
where $x^{\text{img}}$ and $x^{\text{txt}}$ denote visual and text tokens, and $z^{3d}$, $z^{m}$, and $z^{a}$ denote trajectory, intention, and action query/noise tokens. The shared trunk builds multimodal context, while each expert updates only its own token span. Later experts read earlier hidden states through read-only key-value caches, yielding \emph{knowledge insulation}~\cite{driess2025knowledge}: upstream representations remain visible to downstream modules without being overwritten by downstream losses. Concretely, the vision-language/trajectory expert is shielded from MANO and action denoising losses, while the robot-action expert can exploit intention cues without collapsing them into a single entangled latent space. The attention mask in Fig.~\ref{fig:attention} follows the same hierarchy: image and text tokens attend bidirectionally, 3D tokens use causal attention within their span while attending to image-text context, MANO tokens attend causally within the intention span and read the full prefix, and robot-action tokens attend to all preceding modalities with bidirectional refinement inside the action chunk. Compared with prior MoT and latent-bridge models~\cite{beingbeyond2026beingh07,luo2026being}, MoT-HRA grounds the intermediate bridge in reconstructed human hand kinematics.

\paragraph{Vision-Language Expert}
The vision-language expert is initialized from a pretrained multimodal decoder~\cite{steiner2024paligemma} and converts visual-linguistic context into coarse spatial waypoints. For a chunk of length $H$, we append $H$ learnable 3D query tokens, each predicting a discretized waypoint $\tau_h = (b_h^x, b_h^y, b_h^z)$ with each coordinate quantized into $B$ bins. The generation process is autoregressive:
\begin{equation}
p_{\theta_{\tau}}(\tau_{1:H}\mid o, w)
=
\prod_{h=1}^{H}
\prod_{c\in\{x,y,z\}}
p_{\theta_{\tau}}(b_h^c\mid o, w, \tau_{<h}).
\end{equation}
We use bidirectional attention over image-text tokens and causal attention over the 3D span, so each waypoint depends on the scene, instruction, and previous spatial anchors while remaining independent of future trajectory tokens. The training objective is coordinate-wise cross-entropy:
\begin{equation}
\mathcal{L}_{3d}
=
- \sum_{h=1}^{H}\sum_{c\in\{x,y,z\}}
\log p_{\theta_{\tau}}(b_h^c\mid o, w, \tau_{<h}).
\end{equation}
This spatial bottleneck asks the model to localize where the interaction should unfold before predicting hand or robot motion. It improves geometric grounding without tying the representation to a specific embodiment, which is important when transferring priors from human videos to robot control. Compared with regressing embodiment-specific controls directly from pixels, the waypoint representation gives downstream experts a compact spatial scaffold that is easier to supervise from reconstructed human demonstrations.

\paragraph{Intention Expert}
The intention expert maps the spatial plan to a human-motion prior. Conditioned on the shared multimodal context and predicted 3D trajectory, it denoises Gaussian noise into a MANO-style hand sequence via conditional flow matching~\cite{lipman2022flow}. Each hand state is
\begin{equation}
m_h = [w_h, j_h], \qquad
w_h \in \mathbb{R}^{7}, \quad
j_h \in \mathbb{R}^{60},
\end{equation}
where $w_h$ stores wrist translation and quaternion orientation, and $j_h$ stacks quaternion rotations for 15 finger joints. Quaternion targets are sign-canonicalized, and decoded quaternions are renormalized before kinematic use. Let $\epsilon^{m}\sim\mathcal{N}(0,I)$, $t\sim\mathcal{U}(0,1)$, and $x_t^{m}=(1-t)\epsilon^{m}+tm$. The intention expert predicts $v_{\theta_m}(x_t^{m}, t, c^{m})$ with condition $c^{m}=(o,w,\tau_{1:H})$ and is trained with
\begin{equation}
\mathcal{L}_{\text{mano}}
=
\mathbb{E}_{t,\epsilon^{m},m}
\left[
\frac{1}{7H}\sum_{h=1}^{H}
\left\|v_h^{w} - (w_h-\epsilon_h^{w})\right\|_2^2
+
\frac{1}{60H}\sum_{h=1}^{H}
\left\|v_h^{j} - (j_h-\epsilon_h^{j})\right\|_2^2
\right]
\end{equation}
We separate wrist and finger losses so the low-dimensional wrist signal is not dominated by the 60-dimensional joint term. MANO tokens use causal attention within their span, preventing valid early states from attending to future or padded suffix tokens, which otherwise destabilizes convergence. By modeling explicit hand kinematics rather than an unconstrained latent code, the expert turns large-scale human videos into an interpretable intention space aligned with physical interaction~\cite{chen2025flowing,romero2022embodied}.

\paragraph{Fine Expert}
The fine expert maps the intention-aware representation to the final robot action chunk with another conditional flow-matching head. Let $a_{1:H}\in\mathbb{R}^{H\times d_a}$ denote an action chunk for a robot with action dimension $d_a$. Starting from Gaussian noise $\epsilon^{a}\sim\mathcal{N}(0,I)$, we define $x_t^{a}=(1-t)\epsilon^{a}+ta$ and train $v_{\theta_a}(x_t^{a}, t, c^{a})$ with
\begin{equation}
\mathcal{L}_{\text{act}}
=
\mathbb{E}_{t,\epsilon^{a},a}
\left[
\left\|v_{\theta_a}(x_t^{a}, t, c^{a}) - (a-\epsilon^{a})\right\|_2^2
\right].
\end{equation}
The conditioning variable $c^{a}$ includes image-text features, 3D trajectory states, and the latent intention states $h^{\mathrm{int}}_{1:H}$. Thus, the policy uses the latent states produced while denoising the MANO representation rather than imitating decoded hand poses as targets. These states encode temporal coordination, uncertainty, and human motor preferences while leaving the final controls embodiment-specific. Following Fig.~\ref{fig:attention}, action tokens attend to all preceding modalities and use bidirectional attention within the action chunk, enabling joint refinement of temporally coupled controls. This is well suited to chunked robot control, where commands within the horizon should be optimized together after the intention scaffold is established.

\paragraph{Joint Training Strategy}
We train the three experts jointly with a partially supervised multi-task objective:
\begin{equation}
\mathcal{L}
=
\lambda_{3d}\mathcal{L}_{3d}
+
\lambda_m \mathbbm{1}_{m}\mathcal{L}_{\text{mano}}
+
\lambda_a \mathbbm{1}_{a}\mathcal{L}_{\text{act}},
\end{equation}
where $\mathbbm{1}_{m}$ and $\mathbbm{1}_{a}$ indicate whether MANO or robot-action supervision is available. Human-demonstration episodes mainly supervise the trajectory and intention experts, while robot-manipulation episodes supervise the trajectory and fine experts. On robot-only samples, the intention expert still provides latent conditioning through the read-only interface but receives no MANO loss, and downstream action gradients are not propagated into the intention cache. This keeps the human-motion prior stable while the fine expert adapts it to robot control. 
This training regime lets human videos shape the intention manifold while robot data specializes embodiment-specific actuation, enabling downstream robot inference without MANO annotations and direct generation of executable real-robot action.

\section{Experiments}

\subsection{Implementation Details}

We train MoT-HRA jointly on HA-2.2M and the real robot dataset~\cite{bu2025agibot} with 64 NVIDIA Hopper GPUs under PyTorch FSDP2~\cite{zhao2023pytorch}. The HA-2.2M dataset supervises the intention expert through MANO-style human-motion targets, the AgiBot-World dataset supervises the fine expert through robot-action targets, and both datasets supervise the vision-language expert through the 3D trajectory prediction objective. Each GPU processes 32 samples, giving a global batch size of 2048, and the model is optimized for 20{,}000 steps with AdamW and gradient accumulation set to 1. We use a base learning rate of $2.5\times 10^{-5}$, $(\beta_1,\beta_2)=(0.9,0.95)$, $\epsilon=10^{-8}$, weight decay $10^{-10}$, a 1{,}000-step warmup, and cosine decay. The chunk horizon is $H=15$. For MANO generation, we apply classifier-free guidance (CFG)~\cite{ho2022classifier} with scale 6.0. Visual inputs are single RGB observations resized to $224\times224$ with padding and standard data augmentation, and language prompts are tokenized to length 256 with instruction dropout 0.1. Unless otherwise stated, we use EMA, activation checkpointing, and 8 data-loading workers per GPU for memory-efficient large-scale optimization.

\begin{table*}[t]
    \centering
    \small
    \setlength{\tabcolsep}{8pt}
    \renewcommand{\arraystretch}{1.15}
    \caption{Hand motion generation results on egocentric~\cite{grauman2022ego4d} and third-person views~\cite{yang2022oakink}. 
    ADE and DTW are reported in meters, while Rot and Joint-Rot are reported in degrees. 
    Lower is better for all metrics.}
    \label{tab:hand_motion_generation}
    \begin{tabular}{lcccccccc}
        \toprule
        \multirow{2}{*}{Method} & \multicolumn{4}{c}{Egocentric View} & \multicolumn{4}{c}{Third-Person View} \\
        \cmidrule(lr){2-5} \cmidrule(lr){6-9}
         & ADE & DTW & Rot & Joint-Rot & ADE & DTW & Rot & Joint-Rot \\
        \midrule
        Being-H0~\cite{luo2025being} & 0.185 & 0.174 & 38.27 & 44.03 & 0.245 & 0.233 & 44.91 & 49.18 \\
        VITRA~\cite{li2025scalable} & 0.154 & 0.146 & 33.26 & 41.81 & 0.211 & 0.201 & 42.59 & 41.72 \\
        \textbf{Ours} & \textbf{0.136} & \textbf{0.127} & \textbf{28.95} & \textbf{34.16} & \textbf{0.184} & \textbf{0.176} & \textbf{38.47} & \textbf{40.12} \\
        \bottomrule
    \end{tabular}
\end{table*}

\begin{figure}
    \centering
    \includegraphics[width=0.95\linewidth]{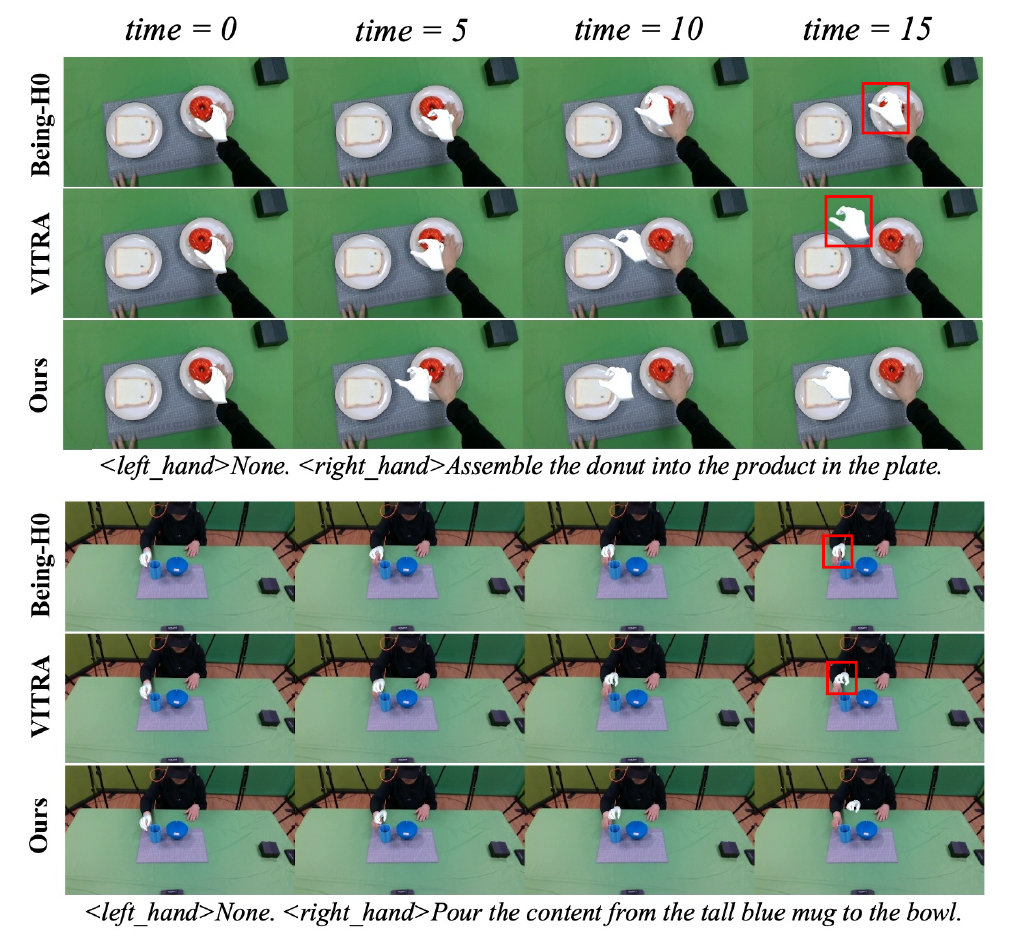}
    \caption{
Qualitative comparison in egocentric (top) and third-person (bottom) views. Our method generates more stable and plausible hand motions. Red boxes highlight typical failure cases of the baselines.}
    \label{fig:mano}
\end{figure}

\subsection{Hand Motion Generation}
We compare MoT-HRA with Being-H0~\cite{luo2025being} and VITRA~\cite{li2025scalable} on held-out egocentric Ego4D~\cite{grauman2022ego4d} and third-person~\cite{yang2022oakink} clips. Although HA-2.2M includes Ego4D videos, the egocentric evaluation split used here is excluded from the training dataset, so it tests held-out egocentric generation rather than memorization of training clips. Third-person view further introduces a larger appearance gap. For each dataset, we evaluate 200 clips, sample five generations per clip with different random seeds, and report average performance. We measure wrist accuracy with average displacement error (ADE, m) and Dynamic Time Warping distance (DTW, m), and pose fidelity with wrist rotation error (Rot, degrees) and finger joint rotation error (Joint-Rot, degrees). Lower is better for all metrics.

As shown in Tab.~\ref{tab:hand_motion_generation}, MoT-HRA performs best across both datasets, improving wrist alignment and articulated pose quality. The advantage on Ego4D indicates that the learned intention prior improves both global motion and local hand configuration, while the gains on third-person view evaluation set suggest that this prior is not limited to the visual style of the training data. The comparatively smaller improvement on fine-grained joint rotation also indicates that articulated finger motion remains the most difficult case under strong domain transfer. Qualitative comparisons in Fig.~\ref{fig:mano} show the same trend: baselines often produce wrist jitter, inconsistent trajectories, or implausible finger poses, whereas our method yields smoother and more anatomically plausible motion. These results support the 3D-to-MANO decomposition, which anchors the interaction spatially before refining it with a structured human-motion prior.

\subsection{Robotic Manipulation}

\paragraph{SimplerEnv Benchmark}
We evaluate MoT-HRA on SimplerEnv~\cite{li2024evaluating} to test visually grounded control under distribution shifts and assess its potential for transfer to real-world manipulation. We focus on WidowX tasks, which vary illumination, background texture, and camera viewpoint, requiring the policy to preserve geometric grounding under distracting observations. This makes the benchmark a stringent test of whether the learned intention representation improves control rather than only fitting human-motion metrics. We compare with a diverse set of strong baselines, including RoboVLMs~\cite{li2026matters}, OpenVLA-OFT~\cite{kim2025fine}, $\pi_0$~\cite{black2024pi_0}, $\pi_0$-FAST~\cite{pertsch2025fast}, SpatialVLA~\cite{qu2025spatialvla}, and ThinkACT~\cite{huang2025thinkact}.

\begin{wraptable}{r}{0.65\textwidth}
    \vspace{-6mm}
    \caption{Success rates (\%) on SimplerEnv-WidowX~\cite{li2024evaluating} tasks. Spoon denotes \emph{Put Spoon on Towel}, Carrot denotes \emph{Put Carrot on Plate}, Stack denotes \emph{Stack Green Block on Yellow Block}, and Eggplant denotes \emph{Put Eggplant in Yellow Basket}.}
    \vspace{2mm}
    \centering
    \scriptsize
    \setlength{\tabcolsep}{3pt}
    \renewcommand{\arraystretch}{1.1}
    \resizebox{\linewidth}{!}{
    \begin{tabular}{lccccc}
        \toprule
        Method & \makecell[c]{Spoon} & \makecell[c]{Carrot} & \makecell[c]{Stack} & \makecell[c]{Eggplant} & Average \\
        \midrule
        RoboVLMs~\cite{li2026matters} & 45.8 & 20.8 & 4.2 & 79.2 & 37.5 \\
        OpenVLA-OFT~\cite{kim2025fine} & 34.2 & 30.0 & 30.0 & 72.5 & 41.7 \\
        $\pi_0$~\cite{black2024pi_0} & 29.1 & 0.0 & 16.6 & 62.5 & 27.1 \\
        $\pi_0$-FAST~\cite{pertsch2025fast} & 29.1 & 21.9 & 10.8 & 66.6 & 32.1 \\
        SpatialVLA~\cite{qu2025spatialvla} & 16.7 & 25.0 & 29.2 & \textbf{100.0} & 42.7 \\
        ThinkACT~\cite{huang2025thinkact} & 58.3 & 37.5 & 8.7 & 70.8 & 43.8 \\
        \midrule
        \textbf{Ours} & \textbf{78.1} & \textbf{62.5} & \textbf{40.6} & 83.3 & \textbf{66.1} \\
        \bottomrule
    \end{tabular}
    }
    \vspace{-4mm}
    \label{tab:simplerenv}
\end{wraptable}

As shown in Tab.~\ref{tab:simplerenv}, our method achieves the best overall performance and the most consistent behavior across tasks. The gains are most visible on \emph{Spoon} and \emph{Carrot}, which require accurate spatial grounding and stable placement under distracting visual changes. This supports the role of the hierarchical intention representation in preserving fine-grained control under distribution shift. Although SpatialVLA is strongest on \emph{Eggplant}, MoT-HRA remains competitive there while avoiding over-specialization to a single scenario.

\begin{figure}
    \centering
    \includegraphics[width=1.0\linewidth]{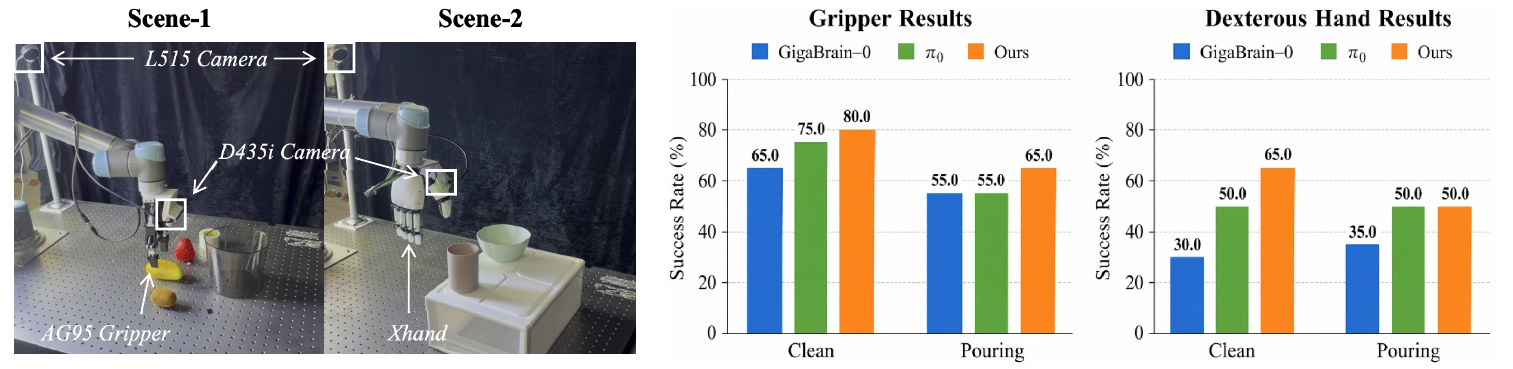}
    \caption{Real-world evaluation on long-horizon manipulation tasks. Both gripper and dexterous-hand settings are tested on \emph{Clean} and \emph{Pouring} under out-of-distribution changes in object position, category, and appearance.
}
    \label{fig:robot}
\end{figure}

\paragraph{Real-World Experiments}
We further evaluate MoT-HRA on real robots with both a parallel gripper and a dexterous hand. Each embodiment is tested on long-horizon \emph{Clean} and \emph{Pouring} tasks after post-training with 150 task-specific trajectories, and evaluation includes out-of-distribution changes in object position, category, and color. Each method is evaluated over 20 trials and compared with $\pi_0$~\cite{black2024pi_0} and GigaBrain-0~\cite{team2025gigabrain}. As shown in Fig.~\ref{fig:robot}, MoT-HRA achieves more reliable task completion across both embodiments, suggesting that the human-intention priors remain effective under realistic visual and object variation.
For additional details, please refer to the supplementary materials.

\subsection{Ablation Study}
As shown in Tab.~\ref{tab:ablation}, the first row can be regarded as a $\pi_0$-style direct VLA architecture, which maps visual-language observations to actions without explicit 3D trajectory grounding, intention experts, or knowledge insulation. Adding the 3D trajectory branch improves both hand-motion metrics and the SimplerEnv average, indicating that spatial grounding provides a useful scaffold for downstream control. Introducing the intention expert brings a larger gain, especially for ADE, DTW, and articulated pose accuracy, showing that explicit human-intention modeling is crucial for coherent wrist trajectories and plausible hand poses.

Knowledge insulation further improves the full model, raising the SimplerEnv average while also reducing motion-generation errors. This suggests that separating intention abstraction from policy learning helps avoid negative interference in large-scale heterogeneous training. Overall, the monotonic improvement from the $\pi_0$-style baseline to the full MoT-HRA model supports the claim that the gains come from the proposed hierarchical structure rather than added capacity alone.

\begin{table*}[t]
    \centering
    \small
    \setlength{\tabcolsep}{8pt}
    \renewcommand{\arraystretch}{1.15}
    \caption{Ablation study of key architectural components. ADE, DTW, Rot, and Joint-Rot are evaluated on Ego4D hand motion generation, where lower is better. SimplerEnv Average reports the average success rate (\%) on SimplerEnv-WidowX, where higher is better.}
    \label{tab:ablation}
    \begin{tabular}{cccccccc}
        \toprule
        \makecell[c]{3D\\Trajectory} & \makecell[c]{Intention\\Expert} & \makecell[c]{Knowledge\\Insulation} & ADE & DTW & Rot & Joint-Rot & \makecell[c]{SimplerEnv\\Average} \\
        \midrule
         &  &  & 0.205 & 0.196 & 40.16 & 44.90 & 48.4 \\
        $\checkmark$ &  &  & 0.182 & 0.173 & 36.64 & 42.53 & 52.1 \\
        $\checkmark$ & $\checkmark$ &  & 0.140 & 0.133 & 30.45 & 35.97 & 62.7 \\
        $\checkmark$ & $\checkmark$ & $\checkmark$ & \textbf{0.136} & \textbf{0.127} & \textbf{28.95} & \textbf{34.16} & \textbf{66.1} \\
        \bottomrule
    \end{tabular}
\end{table*}

\section{Conclusion}
We presented HA-2.2M, a 2.2M-episode action-language dataset reconstructed from human videos, and MoT-HRA, a hierarchical VLA framework for learning human-intention priors. MoT-HRA separates spatial grounding, human-motion modeling, and embodiment-specific action generation, enabling human videos to supervise robot policies without robot actions in every clip. Experiments on hand motion generation, SimplerEnv, and real-world manipulation show better motion plausibility and control robustness, with ablations validating each component. These results suggest that structured human-intention representations can help scale robot policies beyond robot-only data.

\paragraph{Limitations}
MoT-HRA is still limited by the quality and coverage of automatically curated human demonstrations. Noise, ambiguous hand-object contacts, and imperfect action-language alignment may weaken the learned intention prior. Our evaluations cover representative hand-motion and manipulation tasks, but not highly dynamic interactions, multi-object long-horizon planning, or very different embodiments. Future work should improve data verification, expand embodiment coverage, and add failure detection for more reliable open-world deployment.

\bibliography{ref}

\end{document}